\documentclass{article}
\usepackage[hmargin=1in,vmargin=1in]{geometry}
\usepackage{algorithmic}
\usepackage{amsmath}
\usepackage{amssymb}
\usepackage{authblk}
\usepackage{framed}
\usepackage{natbib}

\title{\bf Convolutional Monte Carlo Rollouts in Go}
\author{
  Peter H.~Jin and Kurt Keutzer \\
  {\normalsize \texttt{phj@eecs.berkeley.edu}, \texttt{keutzer@berkeley.edu} }
}
\affil{
  Department of Electrical Engineering and Computer Sciences \\
  University of California, Berkeley
}
\date{}

\begin{document}

\maketitle

\begin{abstract}
In this work, we present a MCTS-based Go-playing program
which uses convolutional networks in all parts.
Our method performs MCTS in batches, explores the Monte Carlo search tree
using Thompson sampling and a convolutional network, and evaluates
convnet-based rollouts on the GPU.
We achieve strong win rates against open source Go programs and attain
competitive results against state of the art convolutional net-based Go-playing
programs.
\end{abstract}


\section{Introduction}

The game of Go remains unsolved by computer algorithms despite advances in the
past decade in Monte Carlo tree search \citep{Kocsis+Szepesvari_2006}.
Recent work in convolutional networks for playing Go
\citep{
  Tian+Zhu_2015,
  Maddison+Huang+Sutskever+Silver_2014,
  Clark+Storkey_2015,
  Sutskever+Nair_2008}
have produced neural move predictors for Go with up to 57.3\% accuracy
on datasets of historical Go game records.
However, in a modern competitive Go computer program, a move predictor is one
component in a Monte Carlo tree search loop.

Previous work used an accurate move predictor built from a very deep
convolutional network to guide the search tree exploration in MCTS
\citep{Tian+Zhu_2015,Maddison+Huang+Sutskever+Silver_2014}.
Exploration is only one half of MCTS;
the other half consists of simulations or rollouts, which in their original form
execute uniformly random moves starting from a leaf node in the search tree
until reaching terminal states to produce fast and unbiased estimates of the
optimal value function for states in the search tree.
Nonuniformly random rollout policies can improve on the winning rate of MCTS
compared to the uniformly random policy.
An early non-uniform rollout policy was pioneered in the Go-playing program MoGo
\citep{Gelly+Wang+Munos+Teytaud_2006}, which matched $3\times3$ board patterns,
first in the local vicinity of the previous move, and then in the rest of the
board.
\citet{Coulom_2007} extended the pattern features by weighting them by relative
strength using a minorization-maximization algorithm and choosing patterns with
probability determined by a Bradley-Terry model.

One reasonable extension is to consider incorporating a convolutional network
in place of traditional pattern-based rollouts.
Because in practice there is a limit to the ``thinking time'' available to
Monte Carlo rollouts, then given a fixed budget of rollouts and a fixed policy
for executing rollouts, it is important to execute as many rollouts in the time
allotted for play as possible.
Simply combining a convolutional net with a sequential MCTS algorithm, such as
UCT \citep{Kocsis+Szepesvari_2006}, is impractical as inference in a convolutional
net has too great a computation latency, even when executed on a high throughput
GPU (ignoring the added communication latency between CPU and GPU)
to perform Monte Carlo backups rapidly enough in MCTS.
Additionally, UCT is deterministic, meaning that its traversal of the tree is
identical in between Monte Carlo backups.
That kind of deterministic behavior is generally undesirable for batched
execution of Monte Carlo rollouts.

Our contribution is threefold:
(1) we implement a MCTS-based Go-playing program that uses convolutional networks
executed on the GPU in all parts;
(2) we perform MCTS in batches to maximize the throughput of convolutions during
rollouts;
and (3) we demonstrate that Thompson sampling \citep{Thompson_1933} during
exploration of the search tree in MCTS is a viable alternative to UCB1
\citep{ucb1}.
Combining those three techniques, we address the earlier concerns,
and our program consistently wins against the open source Go program GNU Go%
\footnote{\texttt{http://www.gnu.org/software/gnugo/}}
and is also competitive against other deep convolutional net-based Go programs.

\section{Related Work}

Convolutional network-based move predictors have been used by themselves for
greedy move selection, or have been applied to the exploration part of MCTS.
\citet{Sutskever+Nair_2008} trained 2-layer convolutional networks for
move prediction in Go.
\citet{Clark+Storkey_2015} and \citet{Maddison+Huang+Sutskever+Silver_2014} later
extended the results to deep convolutional networks.
More recently, \citet{Tian+Zhu_2015} showed that training with multiple labels
for long term prediction further improved the accuracy and playing strength of
deep convolutional move predictors.
In the above works, the rollout part of MCTS, if implemented at all, consisted of
traditional pattern-based rollouts.

Several non-convolutional net-based methods for predicting or ranking moves
have been introduced in the past.
The pioneering Go-playing program MoGo featured pattern-based rollouts
\citep{Gelly+Wang+Munos+Teytaud_2006}.
\citet{Stern+Herbrich+Graepel_2006} learned patterns and local features using
Bayesian ranking.
\citet{Coulom_2007} computed the likelihood of patterns and local features
with a Bradley-Terry model.
\citet{Wistuba+Schmidt-Thieme_2013} used latent factor ranking to achieve move
prediction accuracy of $41\%$.

Thompson sampling has been applied to Monte Carlo tree search in non-computer Go
domains in
\citet{Perick+Maes+Ernst_2012},
\citet{Bai+Wu+Chen_2013},
and \citet{Imagawa+Kaneko_2015}.
In those previous works, Thompson sampling was compared with UCB1 or other
bandit algorithms based on their performance on specific tasks (e.g.,
maximizing reward),
rather than our focus of using Thompson sampling guide batch parallelism.

\citet{parallel-mcts} and \citet{parallel-mcts2} introduced approaches for
parallelizing MCTS, including ``root parallelism'' and ``tree parallelism.''
Like root parallelism, our batching approach traverses the tree multiple
times between consecutive backups;
however, root parallelism does not share backups between parallel trees,
whereas batched backups naturally do.
Unlike tree parallelism which is asynchronous, batching is bulk-synchronous.

\section{Terminology}

First, we clarify some of the terminology that we will use throughout the rest
of this paper.

In the Monte Carlo tree search literature, the terms
\emph{rollout}, \emph{playout}, and \emph{simulation}
are often used synonymously to define the randomly simulated plays.
To avoid confusion, we try to exclusively use the term \emph{rollout}.

In MCTS (see Figure \ref{fig:mcts} for pseudocode),
neural networks can be applied in two different parts:
(1) during exploration of the tree, and (2) during rollouts.
Both of those parts are similar in that, given a state $s$, a classification
neural network can be used to produce the probability of selecting any action
$a$ from the range of available actions at $s$
(for example, when the network has a softmax probability output layer).
Because such a neural network is effectively computing the policy probability
$\pi(s,a)$, we call that network a \emph{policy network}.

More specifically,
we call a neural net used to compute the prior knowledge probabilities of moves
at a fixed game position during exploration of the tree by the term
\emph{prior policy network (PPN)}.
Similarly, a neural net used to compute probabilities of the next move to take
from a single position during a rollout is called a
\emph{rollout policy network (RPN)}.

Using the above terminology, we can briefly compare our method with those of
state of the art convolutional net-based Go programs.
In Table \ref{tab:compare}, we see that our method utilizes both a prior policy
network and a rollout policy network.
On the other hand, previous strong methods use only prior policy networks,
supplementing it with traditional pattern-based Monte Carlo rollouts.

\begin{table}[!ht]
  \begin{center}
    \begin{tabular}{lll}
      Method & Exploration & Rollouts \\
      \hline
      Our method (PPN+RPN)
          & 12 layer PPN, 128 filters & 2- or 3-layer RPN, 16 filters \\
      \citep{Maddison+Huang+Sutskever+Silver_2014}
          & 12 layer PPN, 128 filters & pattern-based \\
      \citep{Tian+Zhu_2015}
          & 12 layer PPN, 384 filters, 1--3 step lookahead & pattern-based \\
    \end{tabular}
  \end{center}
  \caption{
    Comparison of our method with other Go programs that combine
    deep convolutional neural nets with MCTS.
    Our method is characterized by using policy networks 
  }
  \label{tab:compare}
\end{table}

\section{Monte Carlo Tree Search}

The strongest modern Go playing programs all use versions of MCTS.
MCTS builds a search tree of game position nodes where each node keeps track of
Monte Carlo values:
the total number of trials that pass through the node,
and the number of successful trials (wins).
The search tree is updated through three repeating phases:
exploration, rollout, and backup.%
\footnote{
  Note that what we call exploration can further be split into two parts:
  the selection of nodes along a path, and expansion of the search tree.
}

The canonical version of MCTS is UCT \citep{Kocsis+Szepesvari_2006},
where the selection criterion in the exploration phase is determined by the
UCB1 multi-armed bandit algorithm \citep{ucb1}.
For a given node in the MC search tree, denote the total number of trials
through the node's $j$-th child as $n_j$, the number of successful trials as
$w_j$, and the total number of trials through all the node's children as
$n=\sum_jn_j$.
Then during the exploration phase, the UCB1 selection criterion chooses the
next child to traverse by taking the $\arg\max$ in (\ref{eq:ucb1}):
\begin{align}
  \arg\max_j\left[ \frac{w_j}{n_j}+c\sqrt{\frac{\log(n)}{n_j}} \right].
  \label{eq:ucb1}
\end{align}
UCT is described with pseudocode in Figure \ref{fig:mcts}.

\begin{figure}[!ht]
  \begin{framed}
  \begin{algorithmic}
    \STATE // Initialization:
    \STATE initialize the search tree with the current state (game position)
    \STATE
    \WHILE{there is time remaining to search}
      \STATE // Exploration:
      \STATE $s$ $\gets$ the root node of the search tree
      \STATE \texttt{path} $\gets\{s\}$
      \WHILE{$s$ is in the search tree}
        \STATE $a\gets$ use UCB1 to select an action (move) from $s$
        \STATE $s\gets$ apply the action $a$ on $s$
        \STATE \texttt{path} $\gets$ \texttt{path} $\cup$ $s$
      \ENDWHILE
      \STATE add $s$ to the search tree
      \STATE 
      \STATE // Rollout:
      \WHILE{$s$ is not a terminal (end-game) node}
        \STATE $a\gets$ select a random action (move) from $s$
        \STATE $s\gets$ apply the action $a$ on $s$
      \ENDWHILE
      \STATE
      \STATE // Backup:
      \STATE \texttt{score} $\gets$ the final outcome of $s$
      \FORALL{$s'\in$ \texttt{path}}
        \STATE update the Monte Carlo values of $s'$ based on \texttt{score}
      \ENDFOR
    \ENDWHILE
    \STATE
    \STATE // Output:
    \STATE choose the action of the root node which has been selected most by UCB1
  \end{algorithmic}
  \end{framed}
  \caption{
    Pseudocode of Monte Carlo tree search using UCB1 as its bandit algorithm
    (UCT).
  }
  \label{fig:mcts}
\end{figure}

The original rollout policy of MCTS consisted of choosing an action according to
the uniform distribution.
However, it was quickly found that using nonuniformly random rollouts
(for example, based on the probabilities of local features and patterns as in
\citep{Coulom_2007})
resulted in stronger play by MCTS-based Go-playing programs
\citep{Gelly+Wang+Munos+Teytaud_2006}.

In addition to the basics of MCTS,
there are a number of widely used heuristic methods for biasing and pruning
the search, including
RAVE \citep{Gelly+Silver_2007},
progressive bias \citep{progressive-bias},
and progressive widening \citep{Coulom_2007}.
Each heuristic is associated with its own set of hyperparameters, and
achieving strong play generally requires hyperparameter tuning.

\section{Neural Move Prediction}

Recent successful move predictors for Go have been trained using supervised
learning on historical game records.
Typically, the board position is preprocessed into a dense mask of relatively
easy to compute binary and real features, which are then provided as the input
to the convolutional network.
Some common input features include
the board configuration (stone positions),
the ko point if one exists,
the number of chain liberties,
and the stone history or distance since last move.
Using our earlier terminology, a neural network-based move predictor is
a kind of \emph{policy network}.

The most accurate convolutional network architectures for predicting moves in
Go tend to be very deep, with at least a dozen layers
\citep{Tian+Zhu_2015,Maddison+Huang+Sutskever+Silver_2014}.
A typical architecture has a larger sized first convolutional layer, followed
by many $3\times3$ convolutional layers.
The layers also have hundreds of convolution filters, and like most modern
deep convolutional nets use rectified linear units as their nonlinearity
\citep{alexnet}.

MCTS can incorporate a policy network to provide prior knowledge with
equivalent experience into the search tree during exploration
\citep{Gelly+Silver_2007}.
Again using our earlier terminology, a policy network providing prior knowledge
probabilities is a \emph{prior policy network}.
While ideally the prior knowledge is derived from a \emph{move evaluator}
(or a \emph{Q-network})
that computes an approximation to the optimal value of the next position or the
optimal action-value of the move,
directly using the probabilities of the prior policy network is also a
passable heuristic \citep{Ikeda+Viennot_2013}.
Intuitively, directly using the probabilities to bias Monte Carlo values
makes sense when the probablities of moves are closely and positively
correlated with the corresponding optimal action-values.
Additionally, it helps that state values and action-values for Go lie in the
unit interval $[0,1]$, so probabilities and values have comparable range.

\section{Batch Thompson Sampling Tree Search}

In theory, it is also possible to use a policy network to select the
nonuniformly random moves during rollouts.
Such a network can be called a \emph{rollout policy network} and would take the
place of traditional pattern-based rollout policies.
However there are two related obstacles preventing convolutional rollout policy
networks from working effectively:
(1) convolutions are computationally expensive,
and (2) UCB1 is a deterministic algorithm.

When performing inference with a convolutional network, one prefers to evaluate
the input in batches to maximize throughput on hardware platforms such as
modern GPUs.
While batching convolutions is a well known technique and forms the basis of
modern minibatch stochastic gradient methods,
for batching to be effective, the input states need to be sufficiently unique.
If one were to naively explore the Monte Carlo search tree in batches using
UCB1, then many of the states within a batch would be duplicate states.
Asynchronous parallel versions of UCT have also encountered this problem,
getting around it by using heuristic ``virtual losses'' to introduce variance
during tree exploration \citep{parallel-mcts2}.

Instead, we substitute for UCB1 the probabilistic bandit algorithm
Thompson sampling \citep{Thompson_1933} as the search policy in MCTS,
a choice justified by recent empirical evidence \citep{Chapelle+Li_2011},
as well as proofs of its comparable regret bounds to those of UCB1
\citep{Agrawal+Goyal_2012,Kaufmann+Korda+Munos_2012}.
Specifically, we use Thompson sampling with Bernoulli rewards (or beta prior) as
described below in (\ref{eq:thompson_sampling}),
in which the optimal action at each time step is selected by choosing the
$\arg\max$ of the randomly sampled values $q_j$:
\begin{align}
  \arg\max_j q_j~\text{where}~q_j\sim\text{Beta}(w_j+1,n_j-w_j+1).
  \label{eq:thompson_sampling}
\end{align}
We incorporate Thompson sampling into a batched MCTS algorithm, with
pseudocode described in Figure \ref{fig:batch_mcts}.

\begin{figure}[!ht]
  \begin{framed}
  \begin{algorithmic}
    \STATE // Initialization:
    \STATE initialize the search tree with the current state (game position)
    \STATE $B\gets$ batch size
    \STATE
    \WHILE{there is time remaining to search}
      \STATE // Batch-Exploration:
      \FOR{$b=1$ \TO $B$}
        \STATE $s_b$ $\gets$ the root node of the search tree
        \STATE \texttt{path}${}_b$ $\gets\{s_b\}$
        \WHILE{$s_b$ is in the search tree}
          \STATE $a\gets$ use Thompson sampling to select an action (move) from $s_b$
          \STATE $s_b\gets$ apply the action $a$ on $s_b$
          \STATE \texttt{path}${}_b$ $\gets$ \texttt{path} $\cup$ $s_b$
        \ENDWHILE
        \STATE add $s_b$ to the search tree
      \ENDFOR
      \STATE 
      \STATE // Batch-Rollout:
      \WHILE{$s_b$ is not a terminal (end-game) node $\forall~b\in\{1,\ldots,B\}$}
        \STATE $\{a_b\}\gets$ select a set of random actions (moves) for each $s_b$
        \STATE $\{s_b\}\gets$ apply the action $a_b$ on $s_b$ $\forall~b\in\{1,\ldots,B\}$
      \ENDWHILE
      \STATE
      \STATE // Batch-Backup:
      \FOR{$b=1$ \TO $B$}
        \STATE \texttt{score}${}_b$ $\gets$ the final outcome of $s_b$
        \FORALL{$s'\in$ \texttt{path}${}_b$}
          \STATE update the Monte Carlo values of $s'$ based on \texttt{score}${}_b$
        \ENDFOR
      \ENDFOR
    \ENDWHILE
    \STATE
    \STATE // Output:
    \STATE choose the action of the root node which has been selected most by Thompson sampling
  \end{algorithmic}
  \end{framed}
  \caption{
    Pseudocode of batch Monte Carlo tree search using Thompson sampling
    as its bandit algorithm.
  }
  \label{fig:batch_mcts}
\end{figure}

In practice, we execute the game rule logic on the CPU and synchronously run
the convolutional network on the GPU.
Although this incurs communication overhead between main memory and GPU memory,
we believe that splitting the work between the two is optimal, especially on
combined multicore and multi-GPU systems.

\section{Experiments}

\subsection{Data}

We used the GoGoD Winter 2014 dataset of professional Go game records
to train our prior and rollout policies \citep{gogod_w14}.
The dataset consists of 82,609 historical and modern games.
We limited our experiments to a subset of games that satisfied the following
criteria:
$19\times19$ board,
modern (played after 1950),
``standard'' komi ($\text{komi}\in\{2.75,3.75,5.5,6.5\}$),
and no handicap stones.
We did not distinguish between rulesets (most games followed Chinese or Japanese
rules).
Our somewhat strict criteria produced a training set of over 57000 games.

For validation, we used a subset of over 1000 high level games from the
KGS Go Server \citep{kgs_ugo_15}.

\subsection{Architectures}

We used a relatively concise input feature representation in the form of a stack
of $16$ planes of $19\times19$ features each.
For each time step and for each player, we tracked
(1) binary features marking the placement of stones on points, and
(2) one-hot features denoting whether a point belongs to a chain with
$1$, $2$, or $\ge 3$ liberties.
We tracked the last two time steps for a total of $16$ feature planes.

We used a total of 5 different network architectures in our experiments.
These architectures are described in Table \ref{tab:archs}.
They consist of a first layer, followed by repeated inner layers, and completed
by a last layer before a softmax probability output layer;
this is the same architectural pattern used by
\citet{Maddison+Huang+Sutskever+Silver_2014} and \citet{Tian+Zhu_2015}.
Three of them (A, B, and C) are very deep nets and are prior policy networks.
The other two (R-2 and R-3) consist of a shallow net and a minimally deep net,
and are used as rollout policy networks.
All convolution layers are zero-padded with unit stride, and the rectifying
nonlinearity $\sigma_\text{ReLU}(x)=\max(0,x)$ is used at all convolutional
layers except the final one, which feeds into the softmax output instead.

\begin{table}[!ht]
  \begin{center}
    \begin{tabular}{lrlllrr}
      Identifier & Features & First Layer & Inner Layers & Last Layer & Depth & Acc.~\% \\
      \hline
      A   & 16  & $9 \times 9 \times 128$ & $3 \times 3 \times 128 \times 1$  & $3 \times 3 \times 1$ & $3$ layers  & $43.1$ \\
      B   & 16  & $9 \times 9 \times 128$ & $3 \times 3 \times 128 \times 4$  & $3 \times 3 \times 1$ & $6$ layers  & $47.8$ \\
      C   & 16  & $5 \times 5 \times 128$ & $3 \times 3 \times 128 \times 10$ & $3 \times 3 \times 1$ & $12$ layers & $50.3$ \\
      R-2 & 16  & $9 \times 9 \times 16$  & ---                               & $3 \times 3 \times 1$ & $2$ layers  & $34.8$ \\
      R-3 & 16  & $9 \times 9 \times 16$  & $3 \times 3 \times 16 \times 1$   & $3 \times 3 \times 1$ & $3$ layers  & $37.0$ \\
      \hline
      \citep{Maddison+Huang+Sutskever+Silver_2014}
          & 36  & $5 \times 5 \times 128$ & $3 \times 3 \times 128 \times 10$ & $3 \times 3 \times 2$ & $12$ layers & $55.2$ \\
      \citep{Tian+Zhu_2015}
          & 25  & $5 \times 5 \times 92$  & $3 \times 3 \times 384 \times 10$ & $3 \times 3 \times 3$ & $12$ layers & $57.3$ \\
    \end{tabular}
  \end{center}
  \caption{
    The 5 architectures A, B, C, R-2, and R-3 are used in our own
    Go-playing program.
    The other two architectures \citep{Maddison+Huang+Sutskever+Silver_2014}
    and \citep{Tian+Zhu_2015} are state of the art deep convolutional nets for
    predicting moves in Go.
    Architectures A, B, and C are for prior policy networks,
    while architectures R-2 and R-3 are for rollout policy networks.
    All layers are convolutional with dimensions described in the format
    $(\text{conv width})\times(\text{conv height})\times(\text{num filters})$,
    with an optional fourth field denoting the number of layers.
    For our own 5 architectures, the validation accuracy is measured on a
    subset of the KGS dataset.
  }
  \label{tab:archs}
\end{table}

\subsection{Training}

We train our convolutional networks to predict the next move given the current
board position using stochastic gradient descent.
We note that \citet{Tian+Zhu_2015} trained deep convolutional move predictors
to predict the $k$ next moves and found that $k=3$ yielded stronger networks
compared to using single-step labels ($k=1$).

For architectures A and B, we initialized the learning rate at $0.01$ and
annealed the learning rate by a factor of $0.1$ every 2 epochs.
For architecture C, we set the learning rate to $0.05$ and did not tune it.
For A, B, and C, we used zero momentum and zero weight decay.

For architectures R-2 and R-3, we followed a similar training protocol as in
\citep{Sutskever+Nair_2008}.
We initialized the training with learning rate $0.1$, then annealed the learning
rate to $0.01$ after 3200 iterations, running SGD for at least 2 epochs.
We used momentum of $0.9$ and no weight decay.

\subsection{Benchmarking}

We compared the winning rates of our Go program against the open source
programs:
GNU Go version 3.8, a traditional Go program, at level 10;
and Pachi version 11.00 ``Retsugen'' \citep{pachi}, a MCTS program,
with fixed $10^4$ playouts per move and pondering during its opponent's
turn disabled.
The main reason we disable pondering in Pachi is that batched convolutional
rollouts as we implemented them are quite expensive to execute:
typical throughput is between 80 rollout/s to 170 rollout/s when
executing on a single GPU.
Parallelizing rollouts on multiple GPUs can significantly improve the
throughput: for example, a system consisting of $8\times$ high-end NVIDIA Maxwell
GPUs can attain a peak throughput of approximately 1000 rollout/s.

\subsection{Results}

We show the winning rate of different variants of our Go program against
GNU Go, where the variants differed in their combinations of prior policy network
and rollout policy network.
We also tested variants where we did not use MCTS and instead greedily played
the prior policy network's best softmax activation;
these rows are marked with ``none/greedy'' under the ``Rollouts'' column.
Our results are listed in Table \ref{tab:depth_vs_winrate}.
There are at least two interesting observations here.
First, with a 6-layer prior policy network and 2-layer or 3-layer rollout policy
network, we are able to obtain comparable results to 12-layer networks without
MCTS \citep{Tian+Zhu_2015,Maddison+Huang+Sutskever+Silver_2014}.
This by itself is a promising result and shows that sophistication in the
rollout policy can make up for the weaker prior policies.
Second, the winning rate of the variants with the 3-layer rollout policy network
seems to be
less than that of programs with the 2-layer rollout policy network,
despite the 3-layer RPN having a higher accuracy than the 2-layer RPN.
While counterintuitive, this is a known paradox that previous authors of MCTS Go
programs have encountered \citep{Gelly+Silver_2007}.
One possible solution is to use Monte Carlo simulation balancing to de-bias the
the rollout policy \citep{Silver+Tesauro_2009}.
We do not explore this idea further in this paper, and instead show remaining
results using only the 2-layer rollout policy network.

\begin{table}[!ht]
  \begin{center}
    \begin{tabular}{lllr}
      Opponent & Exploration & Rollouts & Win \% \\
      \hline
      GNU Go
      & 3 layer PPN (A) & none/greedy       & $51.0\pm2.5$ \\
      & 3 layer PPN (A) & 2 layer RPN (R-2) & $74.0\pm2.2$ \\
      & 3 layer PPN (A) & 3 layer RPN (R-3) & $66.8\pm3.3$ \\
      & 6 layer PPN (B) & none/greedy       & $90.0\pm1.5$ \\
      & 6 layer PPN (B) & 2 layer RPN (R-2) & $97.0\pm1.2$ \\
      & 6 layer PPN (B) & 3 layer RPN (R-3) & $96.5\pm1.3$ \\
      \hline
      GNU Go
      & 12 layer PPN \citep{Maddison+Huang+Sutskever+Silver_2014}
        & none/greedy & $97.2\pm0.9$ \\
      & 12 layer PPN, 1-step \citep{Tian+Zhu_2015}
        & none/greedy & $96.7\pm1.7$ \\
      & 12 layer PPN, 3-step \citep{Tian+Zhu_2015}
        & pattern-based & $99.7\pm0.3$ \\
    \end{tabular}
  \end{center}
  \caption{
    The effect of network depth on win rate.
    The prior policy networks have width of 128 filters.
    In the rows where MCTS rollouts are enabled, the rollout policy networks
    have width of 16 filters, and the number of rollouts per move is 5120
    with a batch size of 64.
    Win rate is measured versus GNU Go and includes standard error bars.
    The number of trials is between 200 and 400 per row.
  }
  \label{tab:depth_vs_winrate}
\end{table}

In Table \ref{tab:pachi_winrate}, we compare the winning rates of our method
when playing against the MCTS-based program Pachi.
The version of our MCTS-enabled method with a 6-layer prior policy network and
a 2-layer rollout policy network perform comparably
to our 12-layer prior policy network without MCTS,
as well as to the 12-layer network of \citet{Maddison+Huang+Sutskever+Silver_2014}
without MCTS.
Our MCTS Go program with a 12-layer prior policy and 2-layer rollout policy is
comparable to the larger 1-step \emph{darkforest} 12-layer network without MCTS
of \citet{Tian+Zhu_2015}.
Their network was trained using extra features, including ko point, stone
history, and opponent rank, and also has 384 convolution filters compared to
128 filters in our network.
The 3-step \emph{darkfores2} 12-layer network with traditional pattern-based
MCTS rollouts by \citet{Tian+Zhu_2015} performs the best against both GNU Go
and Pachi, although they also employ an enhanced training method
using 3-step lookahead for long-term prediction and tuned learning rate,
whereas we only train our networks using single-step lookahead.

\begin{table}[!ht]
  \begin{center}
    \begin{tabular}{llrlr}
      Opponent & Exploration & Prior Acc.~\% & Rollouts & Win \% \\
      \hline
      Pachi
      & 6 layer PPN (B)   & $47.8$  & none/greedy       & $21.8\pm2.1$ \\
      & 6 layer PPN (B)   & $47.8$  & 2 layer RPN (R-2) & $45.5\pm3.5$ \\
      & 12 layer PPN (C)  & $50.3$  & none/greedy       & $46.0\pm3.5$ \\
      & 12 layer PPN (C)  & $50.3$  & 2 layer RPN (R-2) & $64.4\pm3.5$ \\
      \hline
      Pachi
      & 12 layer PPN \citep{Maddison+Huang+Sutskever+Silver_2014}
        & $55.2$  & none/greedy & $47.4\pm3.7$ \\
      & 12 layer PPN, 1-step \citep{Tian+Zhu_2015}
        & $57.3$  & none/greedy & $67.3$ \\
      & 12 layer PPN, 3-step \citep{Tian+Zhu_2015}
        & $57.3$  & pattern-based & $95.5$ \\
    \end{tabular}
  \end{center}
  \caption{
    Win rate versus GNU Go and Pachi.
    In the rows where MCTS rollouts are enabled (i.e., the rollout depth is
    2 layers), the number of rollouts is 5120
    with a batch size of 64.
    The number of trials is between 200 and 400 per row.
  }
  \label{tab:pachi_winrate}
\end{table}

We explore the effect of different MCTS batch sizes on the overall winning rate
of our program against Pachi, which are shown in
Table \ref{tab:batchsize_winrate}.
Interestingly, varying the batch size between 64 and 256 has little effect on
the winning rate.
This is a very promising result, as it suggests that given a fixed number of
rollouts, increasing the batch size or equivalently decreasing the
number of batch backups up to a limit does not penalize playing strength.

\begin{table}[!ht]
  \begin{center}
    \begin{tabular}{lllrrr}
      Opponent & Exploration & Rollouts & Batch Size & Num.~Backups & Win \% \\
      \hline
      Pachi
      & 12 layer PPN (C)  & 2 layer RPN (R-2) & 64  & 80  & $64.4\pm3.5$ \\
      & 12 layer PPN (C)  & 2 layer RPN (R-2) & 128 & 40  & $66.0\pm3.4$ \\
      & 12 layer PPN (C)  & 2 layer RPN (R-2) & 256 & 20  & $68.8\pm2.9$ \\
    \end{tabular}
  \end{center}
  \caption{
    The effect of varying MCTS batch size on win rate.
    The total number of rollouts per turn is held constant at 5120 for all three
    rows.
    The number of experimental trials is between 200 and 256 per row.
  }
  \label{tab:batchsize_winrate}
\end{table}

\section{Discussion}

In this work, we demonstrated that combining convolutional networks in the
exploration phase of MCTS with convolutional nets in the rollout phase is
practical and effective through Thompson sampling-based batched MCTS.
We evaluated our Go program against the open source programs GNU Go and
Pachi and found that they achieved win rates competitive with the
state of the art among convolutional net-based implementations.
We also found that the winning rate of batched MCTS with convolutional networks
is fairly insensitive to reasonable values of the batch size, suggesting that
further scaling can be done.

While we compared the winning rate of our own Go program against open source
Go programs and the reported results of convolutional net-based Go programs,
there are commercial or otherwise closed source Go programs which are much
stronger:
these include the programs Zen, Crazy Stone \citep{crazystone}, and Dol Baram,
which have won handicap games against highly ranked professional Go players.

Future work includes incorporating batched MCTS with approaches for
training stronger convolutional nets, such as using long-term prediction for
training \citep{Tian+Zhu_2015}
and applying deep reinforcement learning methods to approximate an optimal
value or action-value function.

\section*{Acknowledgments}

Thanks to Forrest Iandola for insightful discussions, and Kostadin Ilov for
assistance with our computing systems.


\bibliographystyle{iclr2016}
\bibliography{draft}

\end{document}